\documentclass[sigconf, screen]{acmart}

\AtBeginDocument{%
  \providecommand\BibTeX{{%
    \normalfont B\kern-0.5em{\scshape i\kern-0.25em b}\kern-0.8em\TeX}}}

\copyrightyear{2021}
\acmYear{2021}
\setcopyright{acmlicensed}\acmConference[FDG'21]{The 16th International Conference on the Foundations of Digital Games (FDG) 2021}{August 3--6, 2021}{Montreal, QC, Canada}
\acmBooktitle{The 16th International Conference on the Foundations of Digital Games (FDG) 2021 (FDG'21), August 3--6, 2021, Montreal, QC, Canada} \acmPrice{15.00}
\acmDOI{10.1145/3472538.3472601} \acmISBN{978-1-4503-8422-3/21/08}

\begin{document}

\title{
Toward Co-creative Dungeon Generation via Transfer Learning
}


\author{Zisen Zhou}
\affiliation{%
  \institution{University of Alberta}
  \streetaddress{116 St & 85 Ave, Edmonton, AB T6G 2R3}
  \city{Edmonton}
  \country{Canada}}
\email{zisen@ualberta.ca}

\author{Matthew Guzdial}
\affiliation{%
  \institution{University of Alberta}
  \streetaddress{116 St & 85 Ave, Edmonton, AB T6G 2R3}
  \city{Edmonton}
  \country{Canada}}
\email{guzdial@ualberta.ca}

\renewcommand{\shortauthors}{Zhou et al.}

\begin{abstract}

\section{Abstract}\label{Abstract}
Co-creative Procedural Content Generation via Machine Learning (PCGML) refers to systems where a PCGML agent and a human work together to produce output content.
One of the limitations of co-creative PCGML is that it requires co-creative training data for a PCGML agent to learn to interact with humans. 
However, acquiring this data is a difficult and time-consuming process. 
In this work, we propose approximating human-AI interaction data and employing transfer learning to adapt learned co-creative knowledge from one game to a different game.
We explore this approach for co-creative Zelda dungeon room generation.
\end{abstract}




\maketitle
\section{Introduction}\label{introduction}

Procedural Content Generation via Machine Learning (PCGML) refers to approaches that employ ML to learn a generative model from existing game content.
PCGML can be applied to co-creative design, where an artificially intelligent (AI) agent collaborates with a human on some game design task.
A significant amount of prior PCGML co-creative work has been applied to Super Mario Bros. (SMB) \cite{guzdial2018co,schrum2020interactive}. 
This is in part due to SMB's popularity.
For example, in one study, a large portion of the participants (62\%) had previously designed a SMB level \cite{guzdial2018co}.
This is likely due to the Mario Maker series, a  Mario level design tool/game. 
The interactions between the study participants and the AI agents from this study were used to train a new AI agent \cite{guzdial2019friend}, which outperformed the original AI agents from the first study.
This would seem to suggest a general process for producing high-quality co-creative PCGML systems: creating a tool with some initial PCGML agents, running a user study, and then using the data from that study to train a new agent.
However, running user subject studies for every game would be costly, and it would be difficult to find a user base with relevant design experience for every game since most games do not have their own \textit{Game Name} Maker level design tool/game.
Therefore, we need a way to develop high quality co-creative agents without requiring game-specific user studies.

Co-creative PCGML agents can be categorized into three groups based on their training data type: hand-authored data \cite{townscaper}, non-interactive or existing game data \cite{schrum2020interactive}, and interaction data \cite{guzdial2019friend}.
Although it is possible to train agents on hand-authored or non-interaction data, evidence suggests training on human interaction data leads to models with a greater positive impact on user experience during co-creative interaction \cite{guzdial2019friend}.
However, as stated above, collecting training data for any desired target game domain is non-trivial. 
One alternative would be to consider reusing the machine learned knowledge from one source game domain for a different target game domain.

Transfer learning is a machine learning methodology focused on transferring knowledge between problem domains.
We could potentially apply transfer learning to adapt machine learned knowledge derived from interaction data for one game to approximate such knowledge for a different game.
However, we would need some interaction data or an approximation of interaction data for the different, target game in order to guide the transfer.
We hypothesize that we can usefully transfer machine learned knowledge from a source game model to a target game model.
If we are able to transfer machine learned knowledge from interaction data between game domains we open up the possibility of producing co-creative Ml agents for any game. 


In this paper, we present an initial exploration of the application of transfer learning for co-creative PCGML agents.
To solve the problem of requiring some interaction data in the target domain we propose three methods to approximate interaction data from non-interaction data.
Given a source game domain model trained on interaction data and non-interactive knowledge for a target game domain, we demonstrate that by finetuning the source domain model, it is possible to positively inherit some knowledge.
We evaluate this approach by comparing this transfer learning-based model to models trained from scratch on the target domain data.

We require a source agent from one game domain and a target agent from a distinct game domain to apply transfer learning. 
We employ the Deep Reinforcement Learning (RL) agent from the Morai Maker tool \cite{guzdial2019friend} as our source agent, as it remains the only example of a co-creative PCGML agent trained on human interaction data.
For a target game domain we chose Zelda dungeon room generation, due to the differences between Mario levels and Zelda dungeon rooms.
We hypothesized that if we could find any evidence of positive transfer between these very different domains, then that would indicate that this could be a general approach to producing co-creative PCGML agents.
By positive transfer, we mean that our transfer approach is able to transfer metaknowledge of interactively working with a human user.
Ideally, when working with a human user in a turn-based fashion, an agent shouldn't make too many or too few additions, as the former cuts off the ability of the human user to impact the results and the latter makes the agent's presence unnecessary.
Thus, if we can find any evidence of this kind of behaviour from transferring the knowledge from the Morai Maker RL agent to a Zelda room design agent, then this would support our hypothesis.

The remainder of this paper is structured as follows. 
Section \ref{related_work} walks through existing approaches and compares them to our approach. 
Section \ref{system_overview} gives an overview of the process needed to apply this approach. 
Section \ref{evaluation} describes the evaluation methods, experiments comparing the transfer learning-based model to models trained from scratch. 
Section \ref{results} gives the results, and Section \ref{discussion} outlines a number of ideas for further improvements.

\section{Related Work}\label{related_work}




In this section we cover related prior work: examples of prior co-creative systems for dungeon design, autonomous dungeon generation, and Reinforcement Learning (RL) for Procedural Content Generation (PCG).

There has been a considerable amount of work in co-creative or mixed-initiative PCG \cite{liapis2016mixed}.
However, there have been relatively few papers on co-creative or mixed-initiative systems for dungeon design, with much of the focus on co-creative systems for 2D platformer level design \cite{guzdial2019friend}.
Alberto et al. \cite{alvarez2018fostering} introduced their ``Evolutionary Dungeon Designer'', using a search-based approach to produce Zelda-like dungeons. 
We employ an ML-based approach instead of a search-based one.
Schrum et al. \cite{schrum2020interactive} use evolutionary search to explore the learned, latent vector of a Generative Adversarial Network (GAN) trained on Zelda dungeon rooms. 
Their approach allowed individuals to explore a latent space of whole Zelda dungeon rooms, whereas our method allows for turn-based, iterative Zelda dungeon room construction.
Depending on how one defines the terms, one can consider the Schrum et al. system ``mixed initiative'' since there is relatively little to no collaboration between the human and AI, whereas our system can be considered ``co-creative'' as the human and AI have more equivalent ability to impacting the output content.
Guzdial et al. \cite{guzdial2019friend} trained their agent on interactive Super Mario Bros. level data collected in an earlier user study \cite{guzdial2018co}. 
We use their model as the source model for our transfer learning approach.

Autonomous dungeon generation has seen a relatively large amount of prior work \cite{viana2019survey}, in comparison to co-creative dungeon generation.
However, there have been relatively fewer PCGML dungeon generation systems.
Summerville et al. used Bayes Networks to capture the distributional information of Zelda level topology \cite{summerville2015learning}, however this work did not generate Zelda dungeon rooms.
Summerville et al. used Principal Component Analysis to extract features from two Zelda rooms, then interpolated between them to produce new rooms \cite{summerville2015sampling}.
Along these lines, Gutierrez et al. trained a Generative Adversarial Network to generate Zelda rooms \cite{gutierrez2020generative}.
All of these approaches could be incorporated into tools to help human users design Zelda dungeons, however they would not allow for the interactive design process of our approach without modification.

PCG via reinforcement learning (PCGRL) has gained academic attention recently \cite{khalifa2020pcgrl}, however it remains under-explored.
Delarosa et al. employed RL agents to give suggestions to human users designing Sokoban levels \cite{delarosa2020mixed}.
We are also interested in mixed-initiative PCGRL, but using a different interactive framework \cite{guzdial2019interaction}, and employing transfer learning. 
A final version of our system would learn while interacting with a human as the source model did \cite{guzdial2019friend}, whereas the agents from the RL Brush tool of Delarosa et al. were all pretrained.
Thue et al. proposed a Procedural Game Adaptation framework that employed RL, that automatically changed the dynamics of a video game during end-user play \cite{thue2012procedural}.
This agent can be viewed as designing the final experience with the player, but the authors do not situate it as a mixed-initiative process or tool.
Nam et al. used PCGRL to generate stages in a turn-based Role Playing Game (RPG) \cite{nam2019generation}.
These and other examples of PCGRL train in hand-authored environments, whereas we attempt to transfer a trained source PCGRL agent's knowledge using existing data.

There has also been work on domain adaptation in games \cite{snodgrass2016approach}.
Snodgrass et al. presented a method of tile-mapping the training data of a source domain to create training data for a target domain.
While utilizing this method could be an approach to transform the interaction data collected from a source game domain to a target game domain, this method only works with game domains with a similar design or genre.
Our approach aims to transfer interaction knowledge across various game domains and is not limited by the design language of a domain.

Our system employs the same interactive framework as Guzdial et al.'s ``Morai Maker'', a turn-based interaction \cite{guzdial2019friend,guzdial2019interaction}.
Many other co-creative or mixed-initiative approaches have instead employed suggestion-based approaches, where a pre-trained or non-ML agent presents variations on existing content that a user can choose between \cite{liapis2016mixed,alvarez2018assessing,delarosa2020mixed}.
In our case, the interaction framework directly impacts how we structure our data for transferring the available knowledge.

\section{System Overview}\label{system_overview}


\begin{figure*}[tb]
    \centering
    \includegraphics[width=\linewidth]{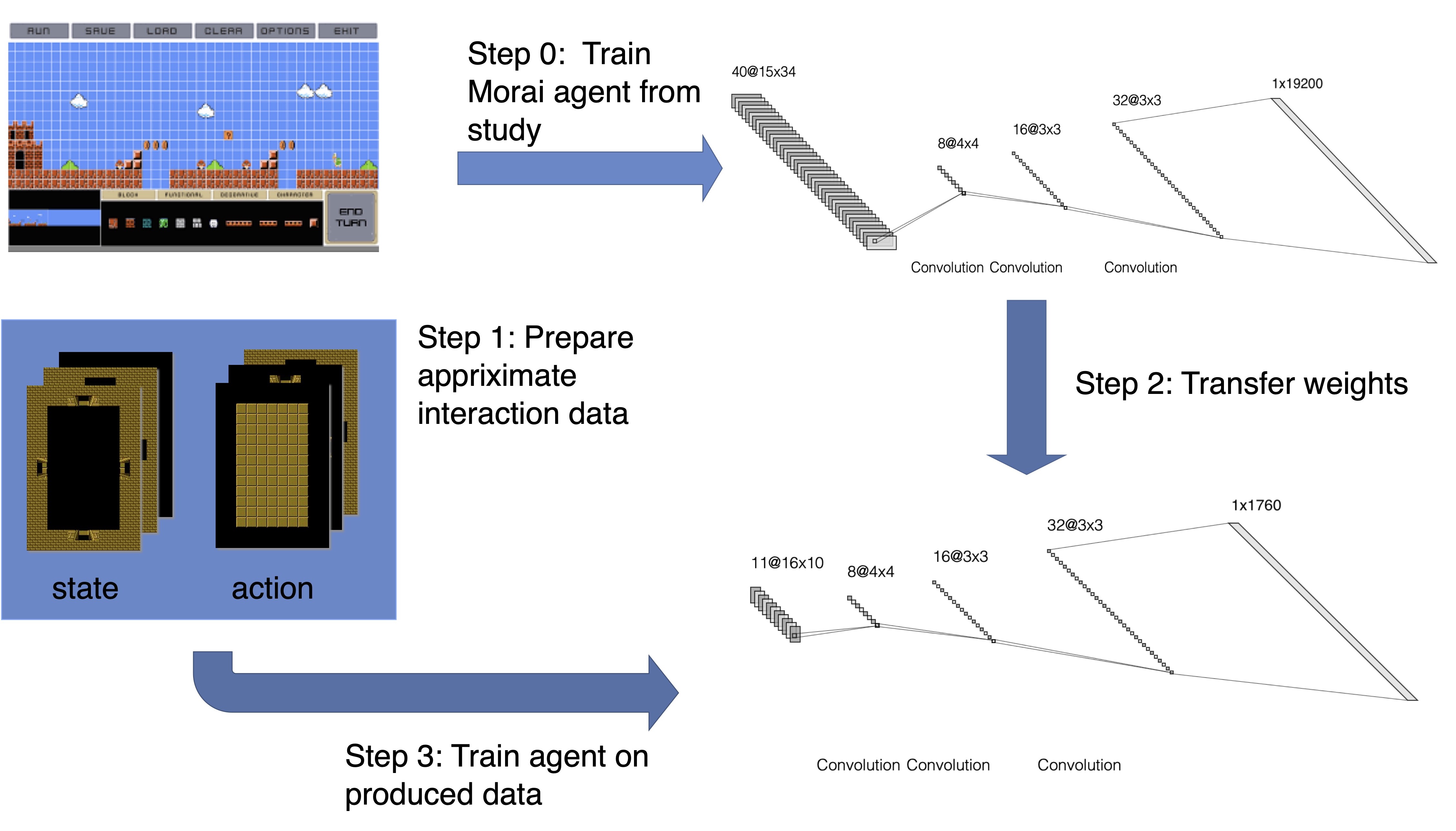}
    \caption{An overview of our transfer learning approach}
    \label{fig:overviewImage}
\end{figure*}

Figure \ref{fig:overviewImage} overviews our approach.
First, we need to have a source model trained on interactions with humans in our source game domain.
In this case, we use Super Mario Bros. as our source game domain, and the Morai Maker deep Rl model as our source model \cite{guzdial2019friend}.
Our goal is to usefully transfer this knowledge to our target model in a different domain.
In our case, our target domain is Zelda dungeon rooms.
We chose this domain as our target domain because it represents a very different type of game level, and we wanted to test whether any knowledge could be usefully transferred between such distinct domains.
Finally, we need a dataset to guide the transfer to the target domain, and it needs to approximate real interaction data in the target domain. 
We explore a number of strategies for approximating this interaction data from non-interactive, complete existing Zelda dungeon rooms. 

\subsection{Source domain}
As discussed above, we used the Morai Maker Deep RL agent model from prior work by Guzdial et al. as our source model \cite{guzdial2019friend}. 
Morai Maker is a level design editor for Super Mario Bros., which is similar to the well-known Mario Maker (literally named because it has ``More AI''), but is research oriented with a co-creative AI partner \ref{fig:overviewImage}.
As a disclaimer, Guzdial et al. did not modify the Super Mario Bros. game released by Nintendo \cite{SMB}, nor modify the hobbyist game ``Infinite Mario Bros.'' \cite{IMB}.
Instead, they produced a Unity version of the game.
The editor includes a large ``End Turn'' button, which switches control between a human editor and AI partner.
The goal of the co-creative AI partner is to be able to make useful additions to the level on its turn, such that the human user will decide to keep the additions.
Because of this, the AI partner typically learned to make a relatively small number of changes, to minimize the chance of deletion, and which allowed the human user to further participate in the design process. 
A number of studies have been run with Morai Maker. 
During Guzdial et al.'s first study they had three Machine Learning (ML) agent options for the AI partner \cite{guzdial2018co}.
The three ML agents were trained on non-interaction data: existing Super Mario Bros. levels.
They ran the first user subject study with the three ML agents and collected data from the interactions between the ML agents and human users.
They then pre-trained a deep RL agent on this interaction data, treating each co-creative level generation process like a rollout from an RL agent. 
This RL agent then further trained on interactions with published game designers in a second user subject study \cite{guzdial2019friend}.
The result for the second user subject study showed that users felt the RL agent adapted to them at a rate of 2 to 1, and the adaptations were well received.
However, desired behaviour for RL agent among users varied widely, with some wanting the agent to do exactly what they did and others preferring more novel behaviour.

Guzdial et al. formulated their co-creative design problem as a semi-Markov Decision Process \cite{rohanimanesh2003learning}, with a state of size 40x15x34.
This represented 40 columns (each 15 tiles in height) of the under-construction level at the point where the user pressed the ``End Turn'' button. 
The 34 represents the Super Mario Bros. entities included in Morai Maker (e.g. ground, bricks, stairs, goombas, koopas, etc.).
Their action size was 40x15x32, representing adding a particular entity (32) at a particular location. 
The reason for the discrepancy (34 vs. 32) was that the human user could place two entities that the agent could not (Mario himself and the end of level flag).
The output of their agent was a Q-table of the same size as the action space, representing the value of adding a particular entity as a particular location.
The reward was defined as follows: for each addition made by the AI partner that the human user kept, the Deep RL agent receives +0.1, and for each addition that the user deletes, the Deep RL agent receives -0.1. 
When the level is completed, the human user gave the Deep RL agent +1 reward if the user would reuse the AI partner and -1 reward if the user would not reuse the AI partner.

\subsection{Zelda Room Generation Markov Decision Process}
In this section, we walk through our Markov Decision Process (MDP) formulation.
We employ the Zelda dungeon room representation from the Video Game Level Corpus (VGLC) \cite{summerville2016vglc}.
This includes the 10 dungeons from the original Legend of Zelda represented in a tile-based representation.
Each room is represented as an 11x16 matrix. 
There are 10 tile types in the VGLC representation, each representing a category of Zelda dungeon entities (e.g. floor, block, monster, etc.). 
Thus, we define our state space $S$ as 11x16x10, or all possible combinations of tiles for an 11x16 room.
We define our action function $A(s)$ as all possible additions of all possible tiles at any location in a state, or 11x16x10 possible actions.
Our Q table values for this MDP then have the same shape, 11x16x10, representing the value of adding a particular tile at a particular location. 
To derive a final policy $\pi$, we employ a threshold $\theta$, and make all additions above $\theta$ for a particular state $s$ in our Q table.
If multiple actions at the same location are above $\theta$, then we take the action with the highest Q value or one at random, depending on the use case. 
We employ the same reward function $R$, as in Guzdial et al.'s Morai Maker work \cite{guzdial2019friend}, with retained additions receiving +0.1, and deleted additions receiving -0.1. 
We also give a +1 reward if a room is considered high quality and -1 reward if a room is considered low quality.
We employ a discount factor $\gamma$ of 0.1, indicating a preference for near term reward. 

\subsection{Model Architecture}
We created our Zelda RL agent such that the architecture is as close as possible to the agent trained by Guzdial et al. \cite{guzdial2019friend}.
There are two reasons for our decision.
First, by having similar layers between source and target models we can transfer as much knowledge as possible.
Second, there is very little data on deep RL architectures for co-creative PCGML, and thus we defer to this existing work.
As shown on \ref{fig:overviewImage}, the neural network layers are exactly the same between the two models, except for the input layer and the output layer.
This is due to the difference in dimensions between the different game level representations.
We employed coefficient of determination as our evaluation metric for development purposes and mean\_square loss as our loss function during training, with all other hyperparameters matching the original work \cite{guzdial2018co}.

\subsection{Data}

We require an approximation of interactive Zelda dungeon room data to guide the transfer learning process.
However, there's no existing data that fits this requirement. 
Even the mixed initiative user study from Schrum et al. \cite{schrum2020interactive} did not produce the necessary training data, since their system produces entire complete levels without human interaction, where we require data from iterative, turn-based interactions.
Thus, we have no choice but to approximate the required interaction data from the non-interactive VGLC data \cite{summerville2016vglc}.

Essentially, we look to approximate a hypothetical situation in which the original human level designers of Zelda designed the rooms with an AI agent partner.
However, there is no literature on how best to approximate interaction data from non-interaction data. 
Thus, we employ three different strategies.
In all cases, we take a final complete room and iteratively remove groups of tiles.
This gives a sequence of states with an initial complete room and a final, empty or nearly empty room (depending on the strategy). 
We can then reverse this sequence to approximate the kind of iterative, incremental design process we require.
The sequence of removals becomes a sequence of additions, a series of actions we treat as coming from the Deep RL agent, despite the fact that the original levels were designed entirely by humans.
We essentially present that instead of the human author creative the original rooms, an AI agent iteratively built each original room with iterative, positive human feedback. 
We have three strategies for determining what tiles are removed/added in each step of this sequence:
\begin{itemize}
  \item \textbf{Tile Type:} For this strategy we iterate through the 10 VGLC tile types. 
  In each iteration all tiles of that type are removed at once. The removal process terminates once all tiles in a room have been removed. 
  We include this strategy because we expect that some users might choose to build a room this way, essentially picking one tile at a time from a palette of options and adding all instances of that tile they desire to a room.
  \item \textbf{3x3:} We randomly select an location as the center of a 3x3 tile section of the room, and remove all tiles that fall within this square. 
  The removal process terminates when three of these 3x3s have been removed. 
  The choice of 3x3 location is random, except that it will not intersect with another 3x3, though they may touch, creating up to 6 tiles wide or tall that have been removed.
  We choose this removal strategy as we think some users may choose to focus on small subsections of a room during the design process.
  \item \textbf{Random:} We randomly remove tiles in a room for this strategy. 
  The removal process consists of multiple iterations, with each step in the process removing 10 tiles at random locations, except when the tile is already empty. 
  The removal process terminates once all tiles in a room have been removed. 
  We choose this removal strategy as an extreme scattershot approach to room generation. 
  We do not expect that users would take this exact approach, but it represents an approach with the least meta-structure for the agent to grasp. 
  In this way it might even be considered a more human-like strategy in terms of the variance of additions. 
  Thus, we anticipate it will be the most difficult type of data to learn from.
\end{itemize}

Each sequence, once reversed, approximates a rollout by the deep RL agent, with each ``reversed'' removal representing an action $a$.
We assign a reward of +0.1 for each $a$, given that every action is ``kept''.
We also give a final reward of +1 at the end of the sequence, because the rooms were already in an existing game, and so we can consider them to be of high quality.
We repeat this process for each removal strategy for each room, giving us three distinct datasets to guide the transfer process.
We separated each room into train and test groups prior to undertaking this process. 
However, each strategy led to different numbers of $s$, $a$, $r$ triplets for training purposes, even though all strategies used the same train and test room sets.  
As such, the Tile dataset had a 544-61 train-test split, the 3x3 dataset had a 1133-126 train-test split, and the Random dataset had a 6820-758 train-test split.

\subsection{Transfer}

We employ a transfer learning approach, which can be considered an example of finetuning or student-teacher transfer learning \cite{zhuang2020comprehensive}.
Thus, we need to transfer the trained weights from each layer of our source network to the corresponding layer of our target network.
For the identical layers, this is a trivial task.
However, for the layers with different shapes we are unable to transfer all available weights. 
In all of these cases, our target model is smaller across all dimensions than the our source model.
For example, our source model's last input dimension is 34, while our last input dimension is 10.
Thus, we transfer only the first 10 of the 34 4x4 filters in the first convolution layer.
While there's no guarantee that the first 10 filters will be the best 10 filters to transfer, we take this strategy throughout all layers with mismatched dimensions. 
Since our source domain and target domain have different design layouts, the knowledge the model will require will also differ.
Therefore, all of the transferred weights are trainable.

\subsection{Training}

The final step of our approach is to finetune the transferred weights on our approximated interaction data.
We employ the adam optimizer as our optimizer due to the variance present in the original Zelda rooms.
We set our batch size to 8 based on initial experiments.
We also set our learning rate experimentally to 0.0001.
For training time, we ran an exhaustive series of tests, training with each dataset from 1-20 epochs, and found that 15 epochs was ideal based on performance on the training data. 
While this number may seem low, it is not unusual for transfer learning to require significantly fewer epochs than training from scratch. 
In addition, we found that training for longer than 15 epochs tended to lead to a loss of any benefit from the transfer weight initialization.

\section{Evaluation}\label{evaluation}
Our method seeks to approximate an interactive model for a game where there is no interaction data available. 
Ideally, we'd use a human subject study to determine the relative benefits of our three interaction data approximations, and whether there was any benefit to employing transfer learning rather than training from scratch even between these very distinct domains. 
However, for this initial exploration of this approach we focus on quantitative experiments. 
The purpose of these experiments will be to give some initial approximation of how well our transfer learning method might work as a co-creative agent.
We employ two baseline agents to compare with our Transfer learning agent. 
Both baselines use the same neural network architecture as our Transfer learning agent, and only differ in terms of the training regimen. 
We call our first agent ``Scratch agent'', which trains on the same approximate interaction data as the Transfer learning agent, but does not transfer knowledge from the source agent. 
This is analogous to a baseline employed in prior work \cite{guzdial2018co}.
Since both the Scratch agent and our Transfer learning agent use our MDP setup, we call them Reinforcement Learning (RL) agents.
The inclusion of this baseline will allow us to investigate whether there's any evidence of positive transfer between the distinct domains of Mario game level design and Zelda room design.

Our second baseline agent is called ``Supervised agent''. 
As the name suggests, it employs Supervised Learning rather than Reinforcement Learning, and it trains on a different yet similarly processed dataset compared to the other two agents.
The Supervised learning (SL) agent directly predicts additions rather than q-table values, using the actual addition as the ground truth.
From each removal strategy described in section \ref{system_overview}, we created two datasets. 
One dataset for the RL agents and one for the SL agent. 
We also name the datasets based on the removal strategy, so we have 6 datasets: two Tile sets, two 3x3 sets, and two Random sets.
Note that the datasets for RL and SL are processed together for each removal strategy, which allows for parallelism between the approaches.
This is important for comparison purposes.
We also remove any duplicates from the datasets to avoid biasing our agents towards overly common patterns like adding walls around the edge of the room. 
We found experimentally that this helps all the agents to add more of the game entities like monsters and blocks.
Otherwise, the agents would just continually predict walls with high confidence.

To evaluate the three agent's performance, we test each agent on a withheld test set for each dataset.
We define two evaluation metrics, which we use to compare the performance of each agent on each test set.
We needed to define these metrics rather than just using a more standard concept of accuracy or an error function like mean square as the RL agents and SL agent were predicting fundamentally different things. 
Our two metrics are:
\begin{enumerate}
    \item \textbf{Action metric:} This metric uses the actual next action/addition for each input state as a ground truth.
    We compare the predicted action from each approach to this ground truth and count up the number of differences between the two.
    Notably, this means that a predicted addition that would later be in the complete room, but was not in the next predefined action would be considered incorrect.
    This metric is meant to approximate how well the agent follows the desired iterative design of the room.
    We can consider each of our three removal strategies to be equivalent to a strict human collaborator, who only wants the AI to make particular additions at each step of the generation process.
    Since we are counting the number of differences we prefer a lower value.
    \item \textbf{Goal metric:} This metric uses the final complete room as a ground truth or ``goal'', and filters out any predicted additions that were already in the input state. 
    In comparison to the Action metric above we are comparing the extent to which the agent predicts additions that would eventually be in the room, since we know ahead of time what all the final rooms should look like in the test set.
    This metric is meant to approximate the what degree the agents attempt to create the final room, even if it ignores the intended, iterative design process.
\end{enumerate}

We require a $\theta$ value for both the RL agents in order to translate their predicted q values into actions/additions. 
While the SL agent does not predict q values, we also require a $\theta$ to translate its output weight activation into additions.
To derive this $\theta$, we have each agent predict on the training sets of each dataset, and select the $\theta$ values that would minimize the Action and Goal metrics over the training set.
We then employ this $\theta$ value when predicting actions for the withheld test set.

We note that high values for the Action and Goal metrics are not necessarily a bad thing. 
Without a human to evaluate the additions, we can only compare to ground truth values (the desired action or desired final room).
However, ``incorrect'' additions may still be useful to a human designer. 
Further, each dataset is only an approximation of how users might iteratively construct Zelda dungeon rooms, but we don't expect any users would follow this strategies this closely in reality. 
Human level designers demonstrate a wide range of level design behaviours \cite{guzdial2019friend}, which these three removal strategies cannot possibly cover. 
However, these still represent three potential level design strategies, and the metrics represent ways human level designers might evaluate their AI partner.

\subsection{Generation}
Given $\theta$ values, we can also generate Zelda dungeon room autonomously. 
For each agent trained on each dataset, we can have the agent predict from an initial empty room, make additions to the room based on $\theta$, then have the agent predict again on the room with previous predictions as input. 
We repeat this process until the room is filled, or there are several blank/void locations that have no predicted values above $\theta$.
In the case where there are multiple tile types for a particular position with values above $\theta$, we randomly choose between them.

We include a third metric to evaluate the agents in terms of their autonomously generated rooms. 
We call this metric \textbf{Output Diversity}, which is meant to indicate how much variance is present between different rooms generated by the same agent. 
For this metric, we simply count up the number of unique tile values at each location for all of the generated rooms. 
To get a sense of this metric over the possibility space of the agent generators, we generate 100 rooms for each agent.
We report the number of differences divided by 100 for this set.

\section{Results}\label{results}

We train three agents for each of the three datasets, meaning we have nine agents in total.
We use a constant random seed across all experiments to ensure they are reproducible and in order to better compare between our agents.
We give the results over the withheld test data for our two metrics over the test data in Table \ref{table:1} and Table \ref{table:3}.

Table \ref{table:1} gives our results for the Action metric. 
These numbers are the average number of differences between the true, desired additions and the predicted additions.
We have bolded the smallest values for each dataset, indicating the smallest average number of differences.
Across all three datasets, the RL agents outperform the SL agent for this metric, with our transfer learning approach outperforming the Scratch agent for two of the three datasets. 
For the Tile set and 3x3 datasets, the RL agents perform very similarly. 
We expect this is due to the fact that both of these datasets had highly structured strategies for removing/adding content. 
As such, there was little benefit to initializing on the human data. 

The Random dataset was unstructured in terms of how tiles were added/removed, and therefore likely to have more human-like variance in its behavior. 
Though the difference might still seem small (~0.2), the Transfer Learning agent outperforms the Scratch agent. 
Note that the random dataset also had the largest test set, meaning this translates to a difference of roughly ~150 mistakes. 
Alternatively, one can consider this as the Transfer learning agent making one fewer mistake in one of every five input test states on average. 
While this is a small improvement, we take it as a positive sign given that the Transfer learning agent had a source domain so distinct from our test domain.
We found that this small ~1 tile improvement was consistent between the Scratch agent and Transfer learning agent qualitatively, which we demonstrate below. 

\begin{table*}
\begin{tabular}{ |l|c|c|c|c|c|c| } 
    \hline
    & Transfer learning agent & Scratch agent & Supervised agent \\
    \hline
    Tile set & \textbf{10.9344}  & 11.0000  & 11.8197 \\ 
    \hline
    3x3 set & 7.2698  & \textbf{7.1429}  & 8.9444  \\ 
    \hline
    random set & \textbf{9.9670}  & 10.1636  & 18.4828  \\ 
    \hline
\end{tabular}
\caption{Action metric results}
\label{table:1}
\end{table*}

\begin{table*}
\begin{tabular}{ |l|c|c|c|c|c|c| } 
    \hline
    & Transfer learning agent & Scratch agent & Supervised agent \\
    \hline
    Tile set & 16.4426 &  16.5574 &  \textbf{15.8361}  \\ 
    \hline
    3x3 set & 5.8968  & 5.9603  & \textbf{0.6667}  \\ 
    \hline
    random set & 68.0752 & 65.8153 & \textbf{33.9472} \\ 
    \hline
\end{tabular}
\caption{Goal metric results}
\label{table:3}
\end{table*}

\begin{table*}
\begin{tabular}{ |l|c|c|c|c|c|c| } 
    \hline
    & Transfer learning agent & Scratch agent & Supervised agent \\
    \hline
    Tile set & 0 &  0 &  0 \\ 
    \hline
    3x3 set & 100.037  & \textbf{100.9184}  & 42.7618  \\ 
    \hline
    random set & \textbf{112.9196} & 110.7303 & 67.0448 \\ 
    \hline
\end{tabular}
\caption{Output diversity results}
\label{table:4}
\end{table*}

Per Table \ref{table:1}, the supervised learning (SL) agent underperformed compared to the RL agents in terms of accurately making the desired additions only.
This is not surprising. 
Given the nature of an iterative design task, supervised learning is less suitable than reinforcement learning. 
The worst performance of the SL agent was on the random dataset. 
We found that performance of the SL agent was due to the agent attempting to just complete the given room, making all additions necessary to do so at every step. 
This led to the worst performance for the Random dataset as it had the longest sequences of additions.

Table \ref{table:3} gives our results for the Goal Metric. 
These values are the average number of incorrect predictions compared to the final output room across each test split for each dataset. 
As such, they are essentially a measure of the degree to which the additions might be accepted by our hypothetical human partner, assuming that human's goal was the final room.
The maximum value of this metric would have been 176, if something were added at every location and it were incorrect. 
Unlike for the Action Metric, the SL agent outperformed both of the RL agents for this metric. 
We anticipate this was due to the fact that the SL agent better generalized over the available training rooms, and so predicted entire rooms that were close to correct for each test instance. 

The Tile set Goal metric values are the closest across all three agents, which we anticipate is due to the similarity of where various tiles are positioned in each room (e.g. walls always occur at the edges of each room). 
The random set performance is especially interesting. 
While the SL agent performed very poorly on this set for the action metric, it clearly outperforms the RL agents for the goal metric.
We found this was due to the SL agent ignoring the desired sequence of additions, and just attempting to fully recreate the room. 
Despite the RL agents poor performance on this metric, the Transfer learning agent slightly outperforms the Scratch agent on two of the three datasets. 

The results from Table \ref{table:1} and \ref{table:3} indicate that reinforcement learning outperforms supervised learning when we want to model iterative design tasks, but not when we only care about the accuracy of a final predicted output. 
They also indicate a slight benefit to employing transfer learning rather than simply training on interaction data directly. 
However, we again note that these metrics cannot fully approximate human evaluation. 
It's possible, as demonstrated in prior work, that agents only trained on existing data may appear highly accurate, but are less able to handle the wide variance of human designer behavior \cite{guzdial2018co}.

Table \ref{table:4} includes our results for the Output Diversity metric. 
The result that immediately stands out is Tile set values, which are 0 across the board. 
This indicates that the agents trained on the Tile training set only ever output a single room each. 
We anticipate this is due to the lack of randomness in the Tile removal method, which only produced a single sequence for each room.
Therefore, generating rooms from an initially empty random input simply led to the same room 100 times. 
In comparison, both the 3x3 and Random sets led to more variance in the output rooms. 
The SL agent performed much worse on this metric than the RL agents, suggesting that it had a tendency to generate fairly similar rooms. 
This suggests again that the SL agent may have generalized more than the RL agents, producing more ``average'' output. 
However, the output from the SL agent may be of higher quality, given the results found in Table \ref{table:3}.
We present further evidence towards this below. 

The Rl agents performed nearly identically for the Output diversity results. 
The Scrach agent had one more different tile on average when trained on the 3x3 dataset, and the Transfer Learning agent had 2 more differences on average when trained on the Random dataset. 
However, we again note that just because an approach output more rooms does not indicate that the rooms were better overall. 
We found that while the Transfer learning agent output fewer unique rooms for the 3x3 dataset that the rooms it did output looked more like the original Zelda rooms.

\subsection{Qualitative Examples}

\begin{figure}[tb]
  \centering
    \includegraphics[width=\columnwidth]{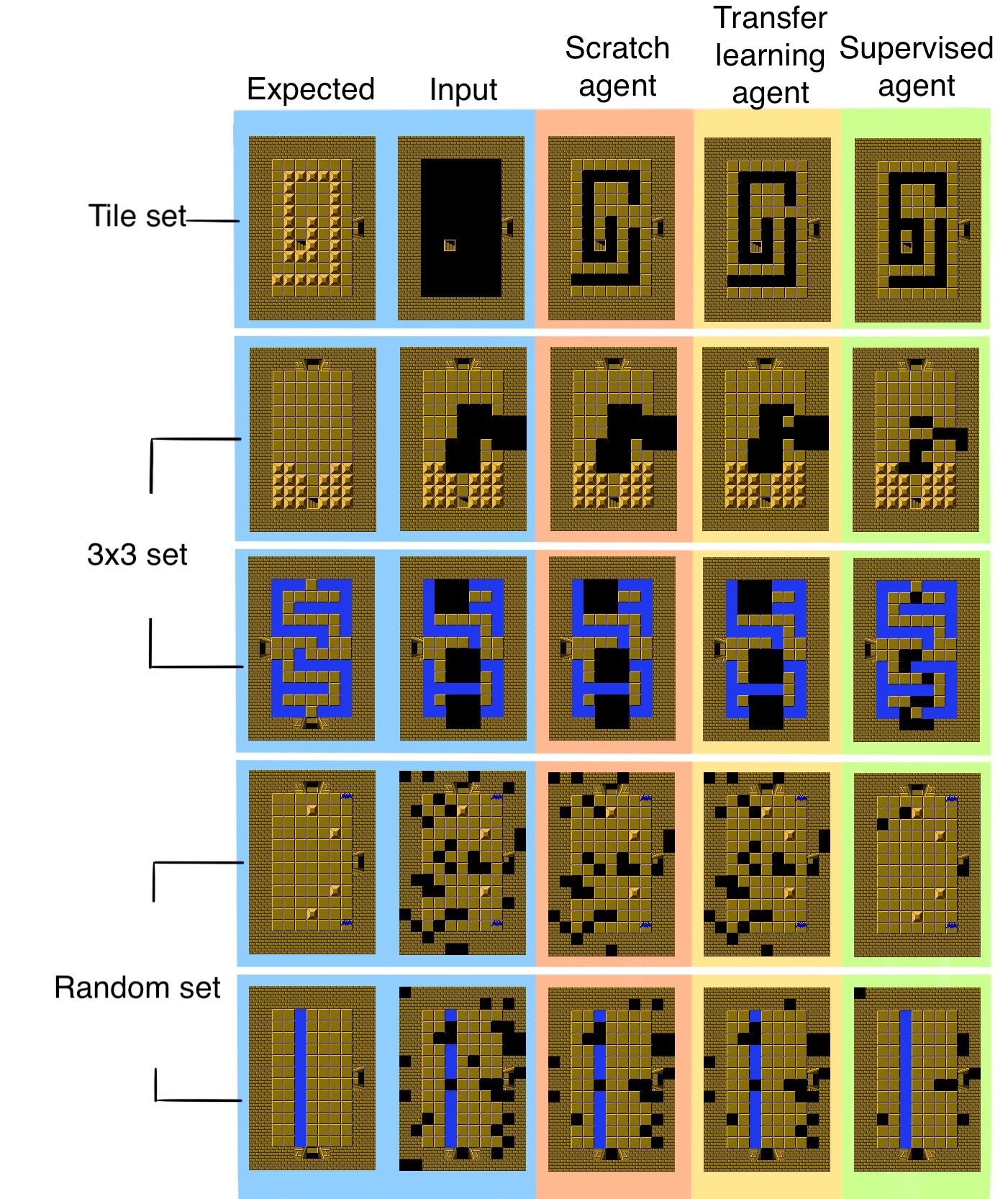}
  \caption{Hand-selected examples of next actions predicted from test instances. }
  \label{fig:predictionComparison}
\end{figure}

We present visualized examples of the next predicted action for each agent and each dataset across a series of hand-picked test examples in Figure \ref{fig:predictionComparison}.
We present one Tile set example and two examples for the other datasets, due to their greater variety.
Each column from left to right indicates the final expected room, the input state to the agent, and then the agent's actions/additions added to the input. 
We identify a number of comparisons between these predicted actions. 
First, the RL agents tended to have only a few differences between them, with only a single difference for the first two rows. 
This matches the quantitative results we reported in the previous subsection.

For the input state in the second row of Figure \ref{fig:predictionComparison}, the Scratch agent could not make any predicted action with the optimal $\theta$ value over the training set.
In comparison, the Transfer learning agent could make one addition.
This may be due to the inherited weights from the source agent, which was forced to always make at least one addition \cite{guzdial2019friend}.
In comparison, the SL agent appears to have done a better job of filling in more tiles that match the final expected room, which parallels the quantitative results found in Table \ref{table:3}.
However, for a turn-based, iterative design task this may not be ideal, since it limits the possibility for the human partner to have an impact on the room's design in their next turn. 
In addition, the actions from the SL agent are not always ideal. 
As we can see in the first row, the SL agent's additions fail to add floor tiles above the staircase, making it impossible for a player to make their way to the staircase from the door. 

\begin{figure}[tb]
  \centering
    \includegraphics[width=\columnwidth]{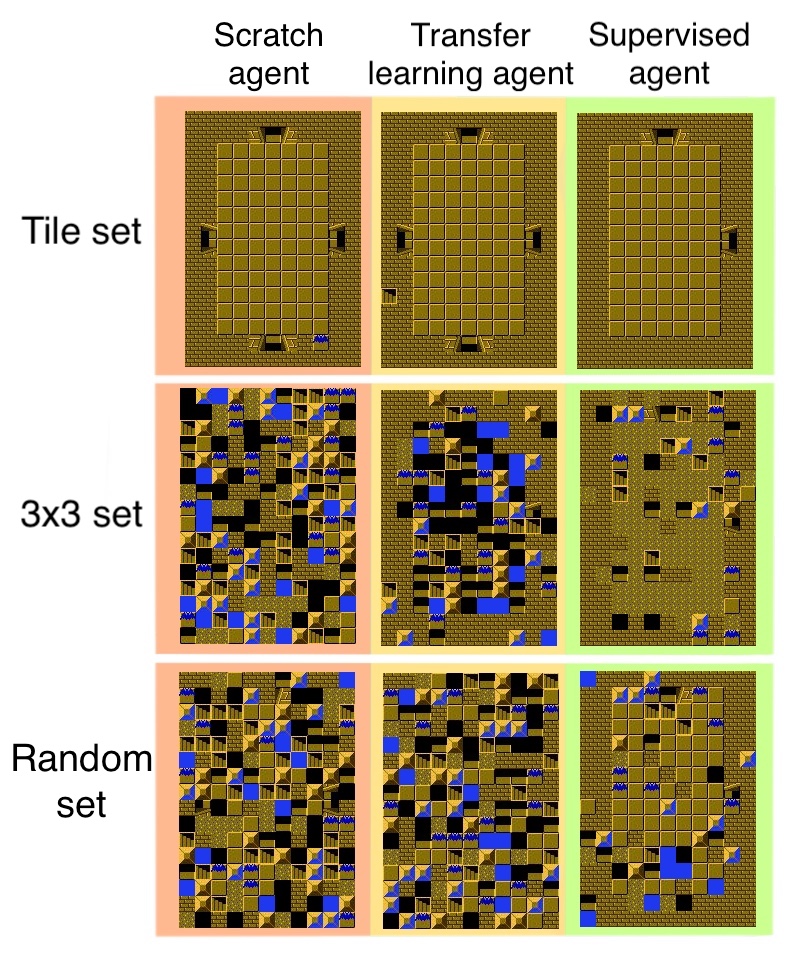}
  \caption{Hand-selected examples of room generation.}
  \label{fig:generationComparison}
\end{figure}

It is more difficult to draw comparisons over the behaviors of the agents for the Random set input states on the final rows of Figure \ref{fig:predictionComparison}. 
For the first row, the Scratch agent made four additions present in the expected final room, compared to the Transfer learning agent's three. 
The Supervised agent, on the other hand, nearly completed the room. 
This is not ideal in the co-creative setting.
While all the Supervised agent's additions were present in the final room, this severely limits a hypothetical human partner in the next ``turn''.
For the last row, the Scratch agent made five additions present in the expected final room, compared to seven from the Transfer Learning agent. 
Taking both of these sets of additions together, this demonstrates the slight edge given to the Transfer learning agent in the quantitative results (9 vs 10 additions between the two rooms). 
Although the difference in the number of additions between the Scratch agent and the Transfer learning agent is small, even a single difference can have a large impact in this domain.
Because a room can only have 176 tiles, the placement of one tile can have a large impact on the design of the room.

The comparison in Figure \ref{fig:predictionComparison} also shows that the Transfer learning agent has a better model of the design of Zelda dungeon rooms compared to the Scratch agent. 
For example, the first row shows that, in the location where there are supposed to be blocks, the Scratch agent twice adds floor tiles where there should be a block, whereas the Transfer learning agent only does this once.
And in the second row, the Transfer learning agent is able to take the first step towards adding a floor whereas the Scratch agent makes no addition.

In Figure \ref{fig:generationComparison} we include one hand-picked example of autonomous room generation from an initially empty room, for each dataset (row) and agent (column). 
We know from the Output Diversity results in Table \ref{table:4} that the rooms present in the first row are the only outputs for each agent trained on the Tile set. 
The only major difference between these outputs are the type of error the RL agents make (one placing an element+block tile in a wall and one placing a staircase in a wall), and that the RL agents create a room with four doors instead of the two doors generated by the SL agent.
The other two datasets led to far more noisy outputs, indicating that the agents trained on these datasets are not well-suited to autonomous room generation from empty inputs. 
For both the 3x3 and Random datasets the SL agent appears to have roughly learned that rooms should have walls surrounding them with some content in the middle, though it has not learned to place floor tiles in the center of rooms when trained on the 3x3 set.
The Transfer learning agent also seems to have roughly learned that rooms are composed of walls with content in the center for the 3x3 set, though it also does not learn to employ floor tiles.
In comparison, on the 3x3 set, the Scratch agent is unable to produce anything but noise.
On the Random set, both agents produce similar noise when attempting autonomous generation. 




\section{Discussion}\label{discussion}



Our purpose for this paper was to investigate transferring the knowledge of an existing co-creative agent for one game domain to a very different game domain. 
We further explored different strategies for approximating interaction data to guide this transfer.
While we indicated previously that we require a human subject study to fully evaluate this work, these initial results do point to some takeaways. 
In particular, our evaluation presents some initial evidence to the tradeoffs between supervised learning and reinforcement learning applied to this task. 
Given these results, we encourage the application of supervised learning to autonomous generation tasks and reinforcement learning to iterative, co-creative tasks. 
While this may seem obvious, we believe these are the first results pointing to this dichotomy for PCGML.

The results of attempting to apply transfer learning to this task are fairly weak. 
While there is evidence that transfer learning is beneficial on certain tasks and for certain metrics, we do not present a strong difference between our transfer learning method and training from scratch. 
However, as indicated above, this may be due to the inability to quantitatively approximate human judgement. 
For example, even in situations where the number of correct additions were similar between the RL agents, it's unclear which of these additions a human partner would prefer.
To more cleanly compare between transfer learning and training from scratch, we would need a human subject study.

One other reason for this lack of clear distinction between training from scratch and transfer learning may be due to how different the source domain (Super Mario Bros. levels) is from the target domain (Zelda dungeon rooms). 
We spoke to this difference above, but one aspect we did not touch on was the difference in terms of the size of the problem. 
The SMB MDP had a state space of 40x15x34, compared to the Zelda MDP state space of 11x16x10.
Given that the state space for the SMB MDP was over ten times larger, this could mean that the Zelda design domain was just too simple to benefit from much of the knowledge present in the source model. 
If so, this could indicate that our decision to take the first $X$ filters/weights for mismatched layers was a poor one.
The reason we ran this initial experiments on this far transfer task was to determine the boundaries of this application of transfer learning.
In future work, we hope to investigate the implications of transferring between more similar source and target domains, and to explore alternative transfer approaches for mismatched weights. 

In the original work our approach is based on \cite{guzdial2018co}, the authors found that training an RL agent on approximated interactive Super Mario Bros. data could outperform an agent trained on interaction data for some tasks. 
However, when compared across a number of human users, the true interaction data agent outperformed the approximated interaction data agent at a ratio of 9:2. 
While this is much better than our 2:1 performance comparing the Transfer learning and Scratch agents, it points to the important of actual interaction data, both for training and testing agents. 
As such, we anticipate that if we had access to some interactive Zelda dungeon room generation data, even if it was not enough to train an agent from scratch, we might be able to use it to better guide the transfer learning process. 
We hope to investigate this further in future work.

\section{Conclusions}\label{conclusion}

In this paper we explored an initial attempt at applying transfer learning to co-creative PCGML. 
We took a source model trained on interactions with humans for Super Mario Bros. level generation and attempted to transfer it to the task of co-creative Zelda dungeon room generation. 
We presented three strategies for approximating interaction data from non-interactive existing data.
Our initial results demonstrated clear takeaways for when to apply Supervised Learning and Reinforcement learning, and some support for applying Transfer learning over training from scratch.
\begin{acks}
This work was funded by the Canada CIFAR AI Chairs Program. We acknowledge the support of the Alberta Machine Intelligence Institute (Amii).
\end{acks}

\bibliographystyle{ACM-Reference-Format}
\bibliography{main}

\end{document}